\documentclass[10pt,twocolumn]{article} 
\usepackage{simpleConference}
\usepackage{times}
\usepackage{graphicx}
\usepackage{amssymb}
\usepackage{url,hyperref}
\usepackage{amsmath}    
\usepackage{booktabs}   
\usepackage{multirow}   
\usepackage{float}      
\usepackage{cite}       
\usepackage{caption} 
\usepackage{titlesec}

\titleformat{\subsubsection}
  {\normalfont\normalsize}      
  {\thesubsubsection.}          
  {0.5em}                       
  {\itshape}                    
\begin{document}

\title{MULTIMODAL SIGNAL PROCESSING FOR THERMAL-VISIBLE-LIDAR FUSION IN
 REAL-TIME 3D SEMANTIC MAPPING}
\author{
    \textit{Jiajun Sun\textsuperscript{1}*, Yangyi Ou\textsuperscript{1}*, Haoyuan Zheng\textsuperscript{2}, Chao Yang\textsuperscript{2}, Yue Ma\textsuperscript{2$\dagger$}} \\[1em]
    \textsuperscript{1}College of Mechatronics and Control Engineering, Shenzhen University, China, 518060 \\
    \textsuperscript{2}School of Robotics, Xi'an-Jiaotong Liverpool University, China, 215123
}

\maketitle

{
    \renewcommand{\thefootnote}{\fnsymbol{footnote}} 
    \footnotetext[1]{These authors contributed equally to this work.} 
    \footnotetext[2]{Corresponding author:Yue.Ma02@xjtlu.edu.cn}                          
}

\thispagestyle{empty}

\begin{abstract}
In complex environments, autonomous robot navigation
and environmental perception pose higher requirements for
SLAM technology. This paper presents a novel method for
semantically enhancing 3D point cloud maps with thermal
information. By first performing pixel-level fusion of visible
and infrared images, the system projects real-time LiDAR
point clouds onto this fused image stream. It then segments
heat source features in the thermal channel to instantly identify high temperature targets and applies this temperature information as a semantic layer on the final 3D map. This approach generates maps that not only have accurate geometry but also possess a critical semantic understanding of the environment, making it highly valuable for specific applications like rapid disaster assessment and industrial preventive maintenance. 
\end{abstract}

\vspace{0em} 

\section{Introduction}
Traditional infrared thermal imaging performs defect identification based on surface temperature distribution, but its detection capability for deep structural defects is severely insufficient\cite{yao2019infrared}. When internal building defects (such as concrete internal voids, rebar corrosion) have not yet caused significant surface temperature differences, 2D thermal imaging is difficult to detect these hidden dangers. Even advanced active thermal imaging technology (AT), when identifying shallow flat-bottom hole defects inside carbon fiber reinforced polymer (CFRP), performs significantly worse than digital shear
speckle technology (DS), which can clearly present subtle
damage boundaries that traditional thermal imaging cannot
distinguish\cite{queiros2021inspection}.
The detection effectiveness of 2D thermal imaging is
highly dependent on environmental conditions, with temperature gradient changes, sunlight intensity, wind speed and other factors significantly affecting detection results \cite{pallares2021structural}. 
In the Miaozhans\`{i} Jingang Pagoda stone carving hollow detection project in Yunnan, research found that only during specific time periods (11:00--12:00) could obvious temperature differences exceeding $7.65^{\circ}$C be observed, while temperature differences in other periods were too low, causing blurred hollow boundaries.

Visible light cameras, although capable of providing high-resolution texture information, have limited performance and can cause image quality degradation in harsh environments such as low light and smoke \cite{zhang2023non}. 
Additionally, traditional methods struggle to provide precise geometric parameters and mechanical state quantification data for defects. 
For example, in stone artifact hollow detection, thermal imaging alone cannot obtain key quantification indicators such as hollow area volume ($6.938 \times 10^5 \text{ mm}^3$ in the study) and deformation height (maximum $13.61 \text{ mm}$) \cite{zhang2023quantifying}.

To address the limitations of traditional 2D imaging, this paper proposes a tri-modal fusion framework for 3D thermal entity reconstruction that integrates visual, spatial, and temporal information. 
The main contributions are: 
(1) a large-scale applicable system combining thermal imaging, localization, and visible light to achieve temperature-texture-geometry joint modeling and 3D semantic point cloud generation; 
(2) a targetless extrinsic calibration method validated against benchmark approaches; and 
(3) field experiments on large structures demonstrating multimodal redundancy for robust performance under sensor failures as shown in Fig.1.

\section{Methods}
\begin{figure}
    \centering
    \includegraphics[width=1\linewidth]{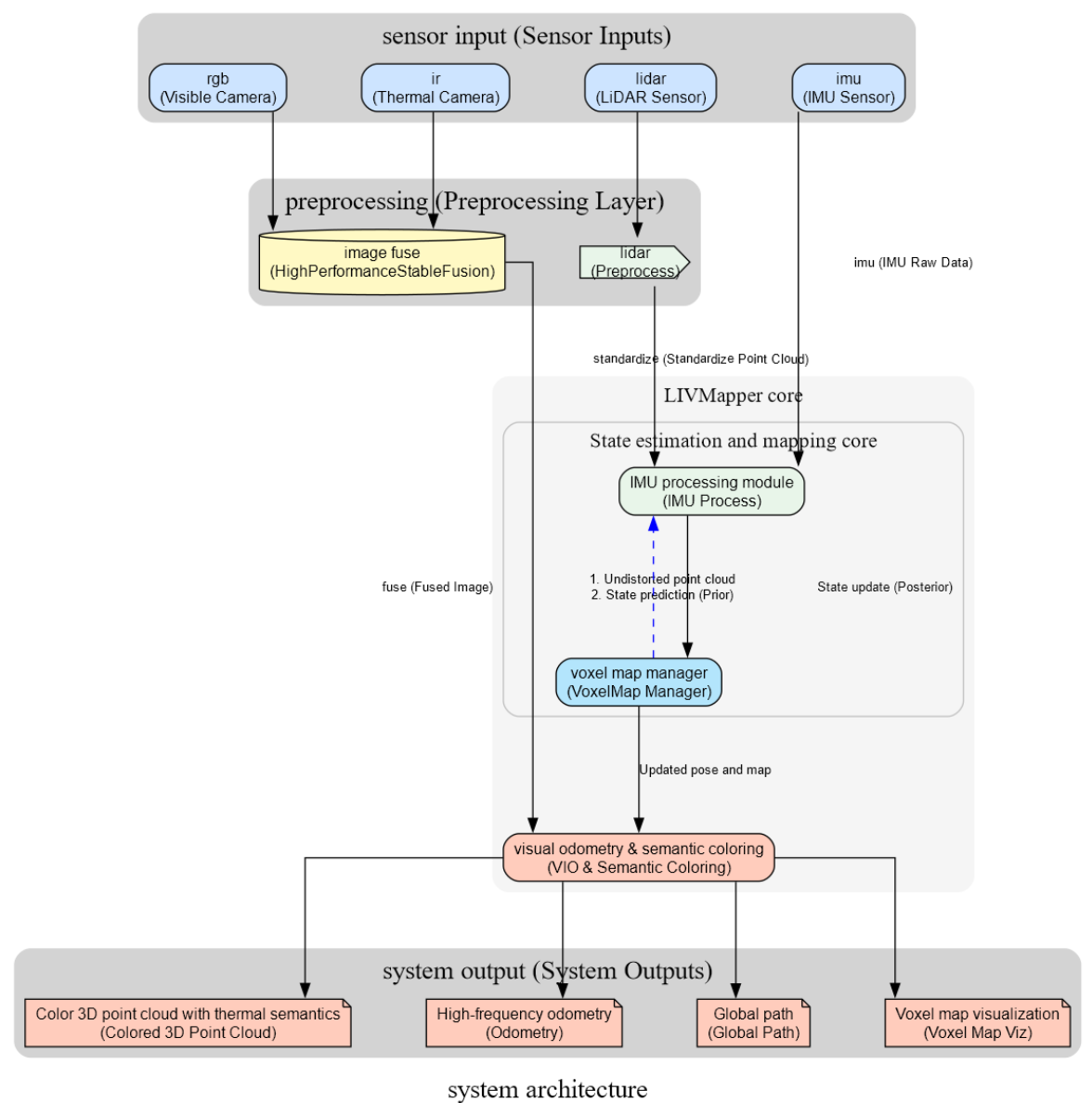}
    \caption{System Framework Overview}
    \label{fig:placeholder}
\end{figure}
The proposed framework uses LiDAR--inertial odometry as the core for high-precision geometry and pose estimation, and introduces a visual--thermal fusion module for real-time semantic enhancement of 3D point clouds. 
LiDAR data are preprocessed with precise timestamps and synchronized with IMU and dual-light cameras; 
IMU pre-integration removes motion-induced distortions, and tightly coupled LiDAR--IMU odometry with scan-to-map matching and octree-based feature extraction achieves accurate state estimation and mapping. 
Thermal and visible images are geometrically aligned, fused into composite textures, and projected onto point clouds, producing RGB maps with both geometric accuracy and physical semantics.

\subsection{System Overall Architecture}
This system integrates three sensors: LiDAR, visible light camera, and thermal infrared camera, constructing a high-precision, multi-information fusion SLAM system. The system architecture is shown in Figure.2.
\begin{figure}[ht]
    \centering
    \includegraphics[width=\linewidth]{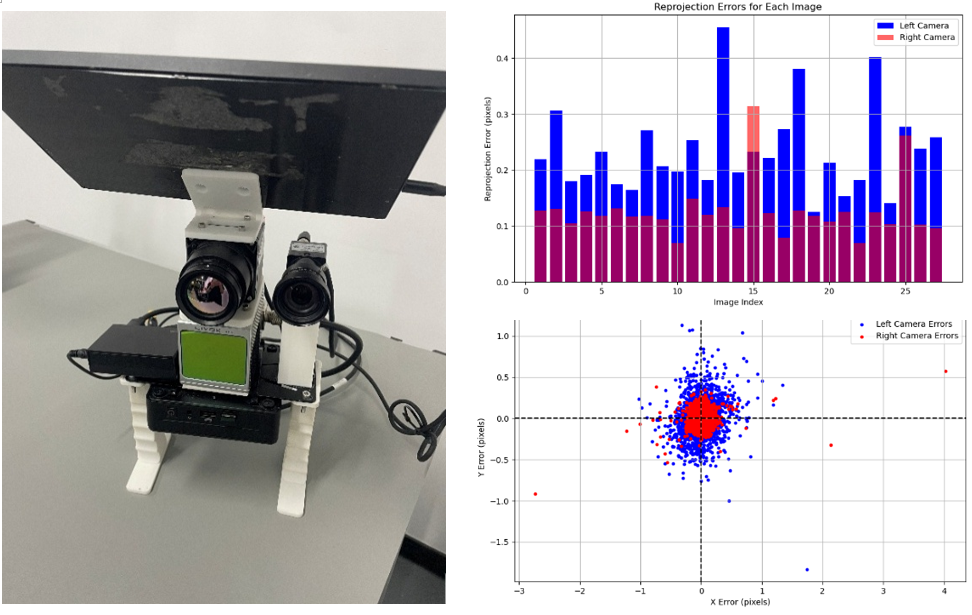}
    \caption{Building 1 (left) and Building 2 (right)}
    \label{fig:buildings}
\end{figure}
\begin{table}[h]
    \centering
    \captionsetup{labelsep=period, labelfont=bf} 
    \caption{Sensor Models}
    \label{tab:sensor_models}
    
    \resizebox{\columnwidth}{!}{%
        \begin{tabular}{ll}
            \toprule
            \textbf{Sensor Type} & \textbf{Model} \\
            \midrule
            RGB Camera          & Hikvision MV-CU013A0UC \\
            Thermal IR Camera   & Hikvision MV-CI003-GL-N6 \\
            IMU                 & Not specified \\
            LiDAR               & Not specified \\
            Synchronization     & Unified clock source \\
            \bottomrule
        \end{tabular}%
    }
\end{table}

\subsection{Multi-sensor Calibration}

\subsubsection{Camera Intrinsic Calibration}
Zhang's calibration method \cite{zhang2000flexible} is used for intrinsic calibration. The camera projection model is:
\begin{equation}
    \begin{bmatrix} u \\ v \\ 1 \end{bmatrix} = 
    \begin{bmatrix} 
        f_x & 0 & c_x \\ 
        0 & f_y & c_y \\ 
        0 & 0 & 1 
    \end{bmatrix} 
    \begin{bmatrix} X_c \\ Y_c \\ Z_c \end{bmatrix}
\end{equation}

\subsubsection{LiDAR-Camera Extrinsic Calibration}

Extrinsic calibration is performed by matching LiDAR edges with corresponding image edges. For $k$-nearest neighbors $\{e_j\}_{j=1}^{k}$, the covariance matrix is calculated as:
\begin{equation}
    C = \frac{1}{k} \sum_{j=1}^{k} e_j e_j^T
\end{equation}

Considering measurement noise, the relationship between true point position and measured value is:
\begin{equation}
    p_i = d_i \cdot r_i = (\hat{d}_i + \xi_d) \cdot \hat{r}_i \exp([B_i \xi_r]_\times)
\end{equation}

\subsection{Infrared-Visible Light Image Fusion}

\subsubsection{Adaptive Thermal Region Detection}
An adaptive threshold strategy is adopted, with the dynamic threshold calculated as:
\begin{equation}
    T_{adaptive} = \text{clip}\left(\frac{\mu + k\sigma - T_{min}}{T_{max} - T_{min}}, 0, 1\right)
\end{equation}
A temporal smoothing strategy is introduced to suppress inter-frame flickering:
\begin{equation}
    I_{smooth} = \alpha \cdot I_{current} + (1-\alpha) \cdot I_{previous}
\end{equation}
The final fused image is obtained through weighted superposition:
\begin{equation}
    I_{fused} = w \cdot I_{thermal} + (1-w) \cdot I_{visible}
\end{equation}

\subsection{State Estimation and Mapping}

\subsubsection{State Vector Definition}
The system state vector is defined as: 
\begin{equation}
    x = [p, v, q, b_{a}, b_{g}, g]^{T}
\end{equation}
where $p$, $v$, and $q$ represent IMU position, velocity, and attitude in the world coordinate system. The IMU kinematics model is:
\begin{equation}
    a^{w} = R(a^{b} - b_{a} - n_{a}) - g
\end{equation}

\subsubsection{Voxelized Mapping}
Adaptive voxel grids are used to construct 3D environment maps \cite{usamentiaga2017infrared}. Measurement residuals are constructed by matching current frame LiDAR points with local planes in the map, and Jacobian matrices with respect to the state vector are calculated for nonlinear optimization.

\section{Experiments and Discussion}

\subsection{Experimental Setup}
As shown in Figure.3, the first object is a university sports plaza with a steel-concrete roof that heats and cools rapidly; emissivity differences between roof and curtain walls cause thermal imaging errors, while high thermal mass slows temperature changes. The second object is a campus hall with cracks and stains, where varying solar angles lead to uneven thermal distribution, limiting single-frame infrared imaging for capturing dynamic evolution.
\begin{figure}[ht]
    \centering
    \includegraphics[width=1\linewidth]{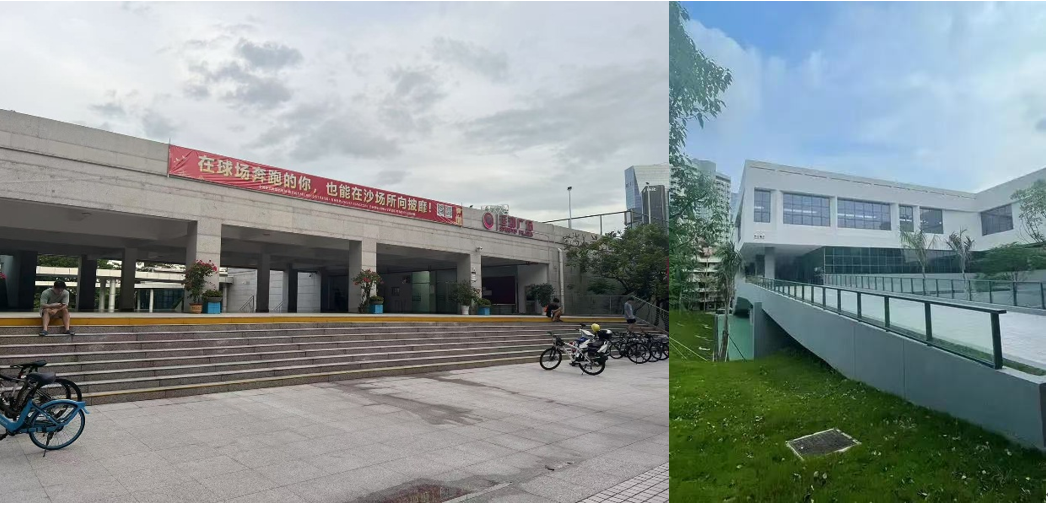} 
    \caption{3D spatial optical-temperature multimodal model of sports plaza. 9:26 (top left), 11:46 (top right), 14:12 (down left), 18:34 (down right).}
    \label{fig:sports_plaza_model}
\end{figure}

Table 2 lists detection parameters for two outdoor structures. The sports plaza and cafeteria walls were scanned at 11:44 and 18:26 under cloudy conditions ($25-30^{\circ}$C) using handheld devices for global modeling. Both mainly consist of concrete, cement, and ceramic tiles with emissivity near 1, so environmental radiation effects were negligible \cite{vermeersch2007fixed}.

\begin{table}[ht]
    \centering
    \label{tab:detection_params}
    \resizebox{\columnwidth}{!}{%
    \begin{tabular}{lcc}
        \toprule
        \textbf{Item} & \textbf{Structure 1} & \textbf{Structure 2} \\
        \midrule
        Structure Type & Sports Plaza & Cafeteria \\
        Location & Univ. Campus & Univ. Campus \\
        Detection Time & 9:00-22:10 & 9:00-22:10 \\
        Detection Area & Front Face & Entire Building \\
        Weather & Clear & Clear \\
        Temp. Range & $25-33^{\circ}$C & $26-30^{\circ}$C \\
        \bottomrule
    \end{tabular}
    }
    \caption{Detection Parameters}
\end{table}

\subsection{Sports Plaza Construction Results}
We collected model data at different times of day, including 11:46 and 18:34, showing that the system can capture dynamic thermal evolution and optical textures of outdoor structures in real time. Clear spatial temperature differences were observed—for example, the sports plaza's shaded right side remained cooler than its sunlit center—demonstrating the strong influence of environmental factors like vegetation, which traditional fixed-angle 2D thermal imaging cannot effectively contextualize \cite{khalid20092d}.
\begin{figure}[ht]
    \centering
    \includegraphics[width=\linewidth]{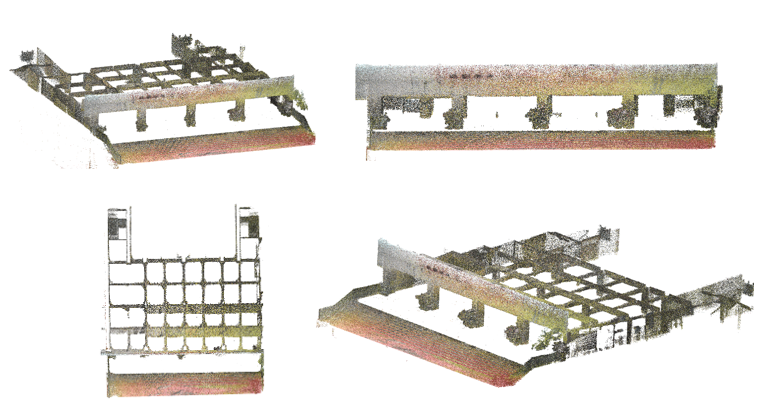} 
    \caption{3D spatial optical-temperature multimodal model of Cafeteria. 9:26 (top left), 11:46 (top right), 14:12 (down left), 18:34 (down right).}
    \label{fig:cafeteria_model}
\end{figure}

The sequence reveals daily thermal cycles, with rising temperatures peaking around noon and cooling after sunset. At 11:46, strong gradients appeared between sunlit and shaded areas, while by 22:10 differences had decreased but residual heat remained, reflecting material response. By fusing thermal and optical data with LiDAR-based 3D models, these patterns can be spatially localized to structural features (e.g., concrete pillars), enabling precise assessment of thermal anomalies and overcoming the limits of traditional 2D imaging for complex or inaccessible areas \cite{rondinella1999materials}.

\begin{figure}
    \centering
    \includegraphics[width=1\linewidth]{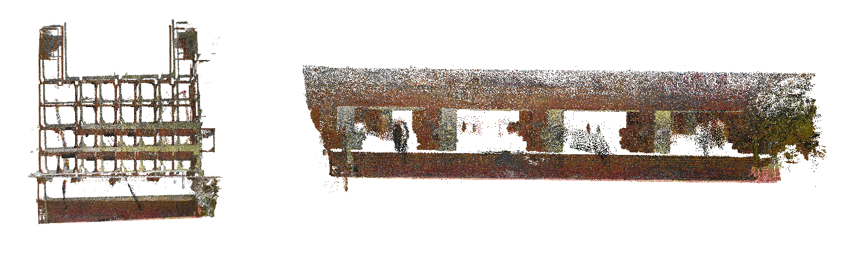}
    \caption{Sports Plaza Model Cross-Sectional View time:11:46
}
    \label{fig:placeholder}
\end{figure}

\subsection{Building Model Construction Results}
Beyond fixed-orientation monitoring results of the sports plaza, Figure \ref{fig:cafeteria_model} shows the system-generated university building 3D optical-thermal multimodal model under the same daily cycle (11:46 and 18:34) as the overpass in Figure \ref{fig:sports_plaza_model}. Results reveal complex spatiotemporal thermal patterns on large building facades under different solar radiation conditions.These models reveal significant temperature difference characteristics between different structural components and materials, such as stark thermal contrast between brown glass doors and surrounding concrete walls.

\begin{figure}
    \centering
    \includegraphics[width=1\linewidth]{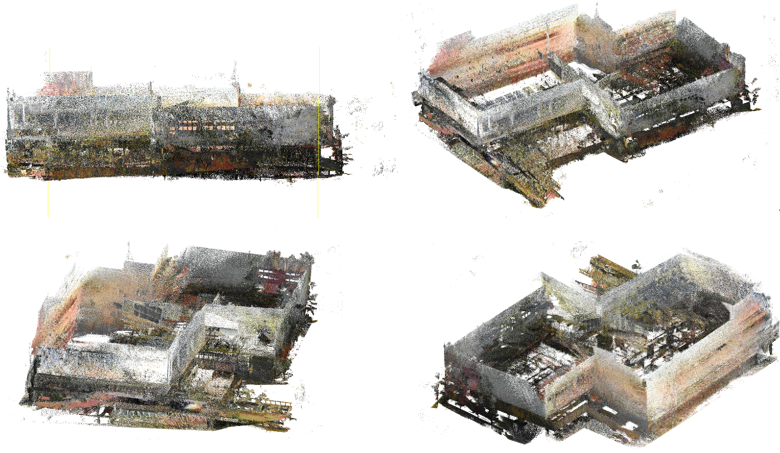}
    \caption{3D spatial optical-temperature multimodal model of
Cafeteria. 9;26(top left) 11:46 (top right) 14:12(down left)
18:34 (down right)}
    \label{fig:placeholder}
\end{figure}

Leveraging fine optical textures, the system accurately locates defects such as cracks, seepage, and stains, while long-term humid areas consistently show lower temperatures that can be mapped onto 3D models. Temporal results confirm dynamic thermal responses under solar loads, with noon heating, lagged peaks due to thermal inertia, and evening cooling \cite{vidas2014real}.

Compared to traditional 3D thermal imaging \cite{shin2019sparse}, integrating thermal data with optical edges addresses dynamic shadows and non-uniform fields, reducing misjudgment in time-varying environments. The Thermal-LIO system further enables precise spatiotemporal localization of anomalies (e.g., persistent low-temperature zones), supporting quantitative assessment and targeted maintenance.

\subsection{Comparative Analysis}
This study validated a multi-perspective thermal diagnostic method that overcomes limitations of single-view imaging by fusing LiDAR geometry with multi-angle visible and thermal data into a unified "Photo-Thermal" 3D model. In a case study, a persistent low-temperature anomaly showed consistent 3D morphology across all views, confirming it as a real defect caused by seepage mud deposition rather than a geometric artifact.

\begin{figure}
    \centering
    \includegraphics[width=1\linewidth]{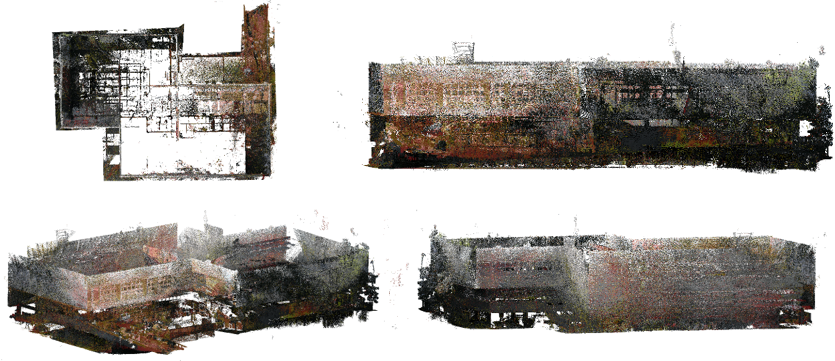}
    \caption{Cafeteria Model Cross-Sectional and Multi
Perspective View time:11:46}
    \label{fig:placeholder}
\end{figure}

\section{Discussion}
This paper presents a multimodal fusion SLAM framework that integrates LiDAR, IMU, and visible-infrared cameras for high-precision mapping and semantic enhancement. Through pixel-level fusion of visible and thermal images and LiDAR projection, the system identifies heat sources and generates 3D maps enriched with geometry, texture, and temperature semantics.

Experiments show stable real-time performance, enabling precise localization of thermal anomalies in 3D space and significantly improving defect diagnosis compared to existing methods. With multimodal redundancy ensuring robustness, the framework offers strong potential for disaster assessment and industrial maintenance. Future work will extend to more complex environments, machine learning-based defect classification, and large-scale infrastructure monitoring.

\begin{table*}[htbp]
\centering
\caption{Comparative Analysis of Key Metrics with Existing Methods}
\label{tab:comparison}
\begin{tabular}{lcccc}
\toprule
\textbf{Metric} & \textbf{Our Method} & \textbf{Shin \& Kim\cite{shin2019sparse} } & \textbf{Chen et al. \cite{chen2021eil}} & \textbf{De Pazzi et al. \cite{depazzi20223d}} \\
\midrule
Frame Rate (FPS) & $>20$ & $<5$ & $8\text{-}12$ & $<3$ \\
Geometric Accuracy & mm-level & cm-level & mm-level & cm-level \\
Detection Range & $>50$ m & $<10$ m & $20\text{-}30$ m & $<15$ m \\
\begin{tabular}[c]{@{}l@{}}Defect Detection\\ Capability\end{tabular} & \begin{tabular}[c]{@{}c@{}}Cracks, seepage,\\ peeling, biological\\ attachment\end{tabular} & \begin{tabular}[c]{@{}c@{}}Structure\\ combination only\end{tabular} & \begin{tabular}[c]{@{}c@{}}Structure\\ + edge\end{tabular} & \begin{tabular}[c]{@{}c@{}}Thermal\\ appearance only\end{tabular} \\
3D Semantic Mapping & Full support & Limited & Partial & Not supported \\
Environmental Adaptability & All-weather & Limited & Moderate & Limited \\
Data Fusion Accuracy (\%) & 94.2 & 78.5 & 85.7 & 76.3 \\
\bottomrule
\end{tabular}
\end{table*}

\section{Conclusion}
This paper introduces a multimodal SLAM framework that combines LiDAR, IMU, and visible-infrared cameras for high-precision 3D mapping with thermal semantics. By fusing visible and thermal imagery and integrating LiDAR data, the system generates detailed maps containing geometric, textural, and temperature information. Experiments confirm robust real-time operation and accurate thermal anomaly detection, greatly improving defect identification.

\bibliographystyle{ieeetr}
\bibliography{refs}

\end{document}